\documentclass[10pt,twocolumn,letterpaper]{article}

\usepackage{cvpr}              %

\usepackage[accsupp]{axessibility}
\usepackage{lineno}
\usepackage{url}
\usepackage{graphicx}
\usepackage{booktabs}
\usepackage{tikz}
\usepackage{pgfplots}
\pgfplotsset{compat=1.17}
\usepackage{multirow}
\usepackage{wrapfig} 
\usepackage{algorithm}
\usepackage[ruled,vlined,linesnumbered,algo2e]{algorithm2e}
\usepackage{array}
\usepackage{soul}      %
\usepackage{xcolor}
\setulcolor{red}       %
\setul{1pt}{.4pt}      %
\usepackage{adjustbox}

\usepackage{amssymb}

\definecolor{cvprblue}{rgb}{0.21,0.49,0.74}
\usepackage[pagebackref,breaklinks,colorlinks,allcolors=cvprblue]{hyperref}
\usepackage{pifont}
\renewcommand{\arraystretch}{1.05}
\usepackage{pifont}
\usepackage{booktabs}
\newcommand{\cmark}{\ding{51}}%
\newcommand{\xmark}{\ding{55}}%
\def\paperID{}%
\def\confName{CVPR}
\def\confYear{2026}

\title{Learning to Reason: Targeted Knowledge Discovery and Fuzzy Logic Update for Robust Image Recognition}

\author{
Gurucharan Srinivas\thanks{Equal contribution.}$^*$\quad
Joshua Niemeijer$^*$\quad
Frank K{\"o}ster\\
German Aerospace Center (DLR)\\
{\tt\small \{Gurucharan.Srinivas, Joshua.Niemeijer, Frank.Koester\}@dlr.de}
}

\begin{document}
\maketitle
\begin{abstract}
Integrating domain knowledge into deep neural networks is a promising way to improve generalization. 
Existing methods either encode prior knowledge in the loss function or apply post-processing modules, but both depend on identifying useful symbolic knowledge to integrate. 
Since such rules are often unavailable in real-world vision tasks, we propose a method for targeted knowledge discovery. 

We propose a Differentiable Knowledge Unit (DKU) that enables modulating the classifier logits, yielding refined class probabilities.
The DKU uses implication rules to represent relationships between task classes and implicit concepts learned entirely from the main task supervision, without requiring concept labels. 
Concepts are identified by dedicated classifiers, whose probabilities are passed to DKU alongside the primary class probabilities. 
DKU computes a logic-based adjustment vector via fuzzy inference, which modulates the primary class logits to yield refined class probabilities.
When concept classifiers represent concepts that do not support the logical rule structure, the resulting adjustments to the class probabilities do not directly minimize the supervision loss.
Consequently, optimizing the supervision loss on these adjusted class probabilities implicitly trains the concept classifiers.
We construct the rule base so that bidirectional logical relations connect concepts and classes.
We enforce the concepts to be distinct from each other and with respect to the classes. 
This design enforces a clean supervision signal for concept learning.

We evaluate our methods on the PASCAL-VOC, COCO, and MedMNIST datasets. 
We demonstrate improvement through our knowledge integration across these datasets. 
We conduct domain generalization and hard-sample ablation studies and find that our implicit knowledge discovery and integration outperforms the baseline.
Code is available: \url{https://github.com/DLR-TS/KLUE.git} 
\end{abstract}

\section{Introduction}
\begin{figure}[t]
    \centering
    \includegraphics[width=1\linewidth]{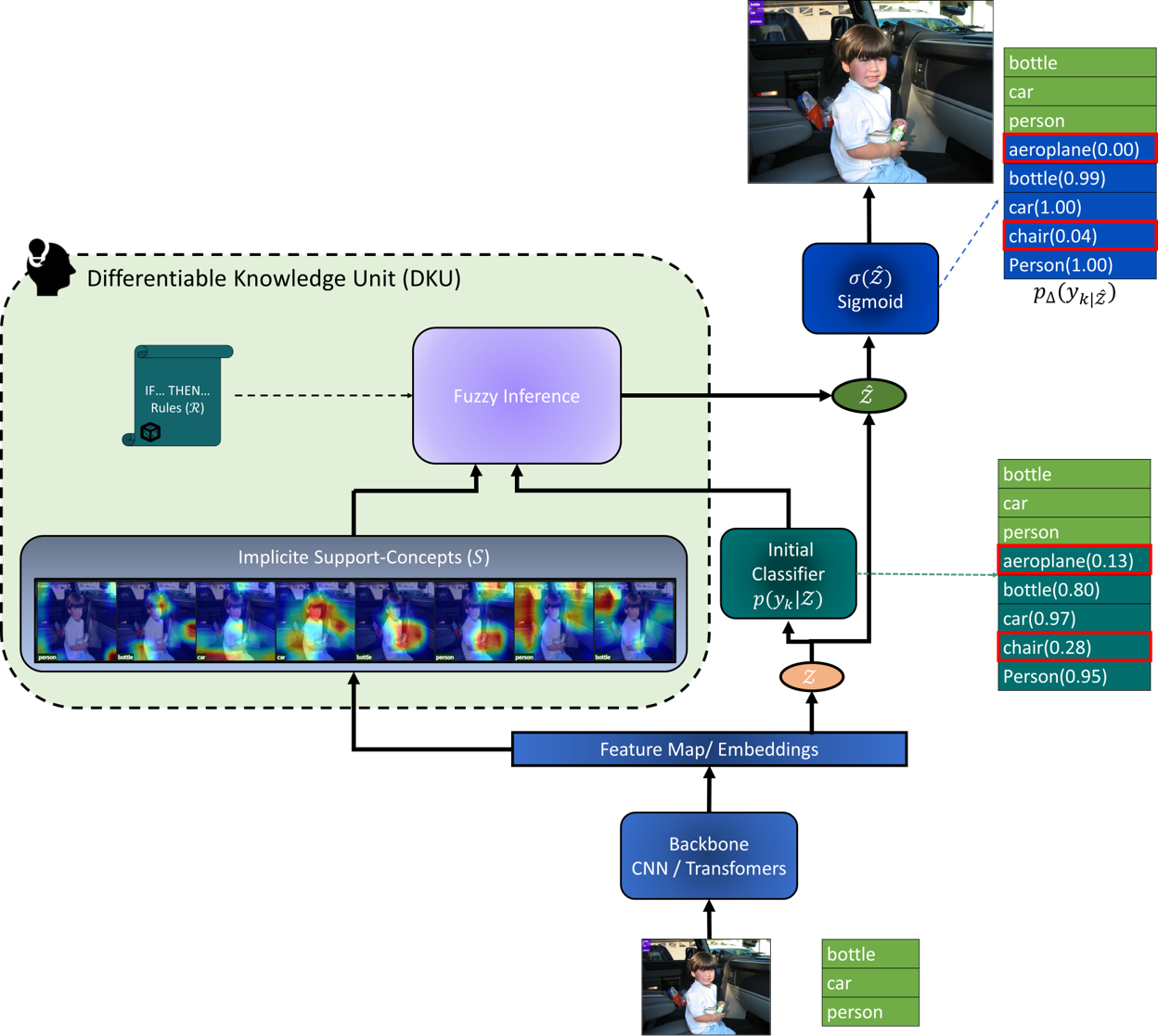}
    \caption{
    The figure shows the impact of integrating self-discovered implicit concepts via our Differentiable Knowledge Unit. 
    Shown: input image, ground truth, initial probabilities $p(y_k|z)$, refined probabilities $p_\Delta(y_k|\hat{z})$, and visualizations of concept activations to provide insight into the learned semantics.}
    \label{fig:simple}
\end{figure}

Deep Neural Networks (DNNs) for perception have enabled the training of discriminative models that yield high accuracy. 
They are trained data-driven and, in the process, optimize encoders that extract image features relevant for classification. 
Network architectures are typically black-box functions, which sets them apart from traditional computer vision systems that rely on human feature engineering.
Such handcrafted systems were therefore more explanatory and often more robust, even though their overall performance was lower.
Neural-Symbolic (NeSy) approaches aim to combine the advantages of data-driven DNNs and the explainability/robustness of handcrafted systems. 
Such approaches try to integrate prior knowledge into DNNs.
There are two major approaches for this integration. 
Either the predicted class probabilities are post-processed using known logical relationships to detect inconsistencies or correlations, or a loss function is designed to embed such relations into the encoder's weights during training.
A typical example of the latter is physics-informed networks, in which a DNN learns physical rules to predict trajectories. 
In discriminative perception tasks, however, it is much less obvious what knowledge matters to the network. 
The fact that acquiring and integrating such knowledge is challenging is likely why feature engineering
did not perform as well. 

The motivation for our work, therefore, is the targeted discovery of knowledge that matters for the neural network.
Therefore, we aim to enable the network to discover task-optimal implicit knowledge referred to as support concepts. 
We define implication rules between several unique support concepts and the target classes.
The DKU encodes these implication rules using fuzzy inference mechanisms to update the class probabilities.
However, since no training labels are available for these support concepts, the concept classifiers must be learned implicitly. 
The implication rules must meaningfully update the class probabilities in order to minimize the overall classification loss.
Thus, the support concepts forming the rules’ elements must encode meaningful information, enabling their implicit training via the standard classification loss.
To avoid trivial self-implications (e.g., a class implying itself), we introduce uniqueness constraints ensuring that each concept remains distinct from its class and from other concepts.
We further impose a bidirectional correspondence between each class and the set of support concepts that define it.

We test the resulting classification network on several datasets: Pascal-VOC 2012, COCO 2014, and the ChestMNIST from the MedMNIST collection. 
We demonstrate improvements over the baseline networks, even with the strong transformer architectures. 
Additionally, we present ablation studies that demonstrate improved performance in corner cases and across domains, as well as enhanced resistance to overfitting. 
We therefore introduce the first approach to automatic, implicit targeted knowledge discovery for neuro-symbolic DNNs. 
We refer to this method as KLUE (Knowledge and Logic Update for Enhanced Recognition).

\section{Related Work}
\textbf{Rule-based and Knowledge-Integrated Deep Learning:}
Integrating symbolic knowledge into neural networks has long been pursued to improve interpretability and incorporate domain expertise. 
Early work, such as KBANN ~\cite{towell1994knowledge}, demonstrated how domain theories represented as propositional logical rules can be mapped into the initial architecture and weights of a neural network.
Furthermore, researchers have developed frameworks to inject logical constraints during training.
Hu \textit{et al.}~\cite{hu2020harnessingdeepneuralnetworks} introduced an iterative distillation method to transfer first-order logic rules into neural network weights, effectively using a teacher-student model to enforce rule satisfaction. 
Whereas Xu \textit{et al.}~\cite{semantic-loss} proposed a semantic loss function 
that measures violation of logical constraints, which can be added to training objectives. 
More recently, Serafini \textit{et al.}~\cite{ltn2016} and Badreddine \textit{et al.}~\cite{LTN2022} proposed Logic Tensor Networks (LTNs), which integrate first-order logic with real-valued neural computation by grounding logical symbols in tensors and fuzzy logic to enable end-to-end differentiable learning that incorporates logical constraints. 
In contrast to LTN, our work shares the goal of integrating logic with deep learning, but differs in the mechanism. 
Instead of a separate loss or distillation process, we embed the rules directly in the model as a differentiable module. 
This is related to LNN~\cite{riegel2020logicalneuralnetworks},
which construct networks corresponding to logic formulas with learnable gates, though our formulation is a fuzzy-logic instantiation for multi-label tasks.
Our approach is also related to concept bottleneck models (CBMs) and attribute-based classifiers \cite{conceptbottleneck}. 
In concept bottleneck models, a network first predicts a set of human-interpretable concept labels from the input, then uses those predicted concepts to predict the final task labels. Our model also includes an intermediate concept-prediction layer; unlike standard concept bottlenecks that rely on ground-truth concept annotations, our approach learns concepts implicitly via logical rules. 
Supervision is applied only to the final labels, while the rules guide the concept outputs to align with meaningful visual patterns.
While Li \textit{et al.} \cite{li2023logicseg} proposed LogicSeg, which uses structured hierarchical class labels to define constraints grounded in the data and network and facilitate logic-induced training, our approach does not rely on labeled hierarchical information. Furthermore, a Neuro-Symbolic computing (NeSy) paradigm for data and knowledge integration has been studied and pursued in \cite{10721277,garcez2019neural,lin2025neurosymbolicspatialreasoningsegmentation,manhaeve2018deepproblog}.

\textbf{Methods of Discovering Knowledge:}
To utilize vanilla variants of LTN or the CBMs, access to domain knowledge is essential for creating a closed rule set \cite{LTN2022} or having labeled concepts and attributes \cite{conceptbottleneck, fasterLTN}, respectively. 
In scenarios where domain knowledge is unavailable or difficult to define, as in common vision tasks, association rule mining algorithms \cite{Apriori,FPGrowth, MSApriori,ELCAT} can be used to derive rules of the form $A\implies C$.
However, applying these methods to large-scale datasets poses significant challenges due to their high computational complexity and data noise, and they do not guarantee the discovery of a closed set of rules. 

Nevertheless, to compare with KLUE, we employ the rule mining algorithm proposed in \cite{FPGrowth} to extract rules and use them within the LTN framework.
Such knowledge discovery, however, is untargeted, meaning there is no optimization incentive to ensure the generated rules are meaningful for the classification task. 
KLUE, on the other hand, directly optimizes the rules to improve classification.

\section{Methods}

\begin{figure*}[t]%
  \centering
  \includegraphics[width=\textwidth]{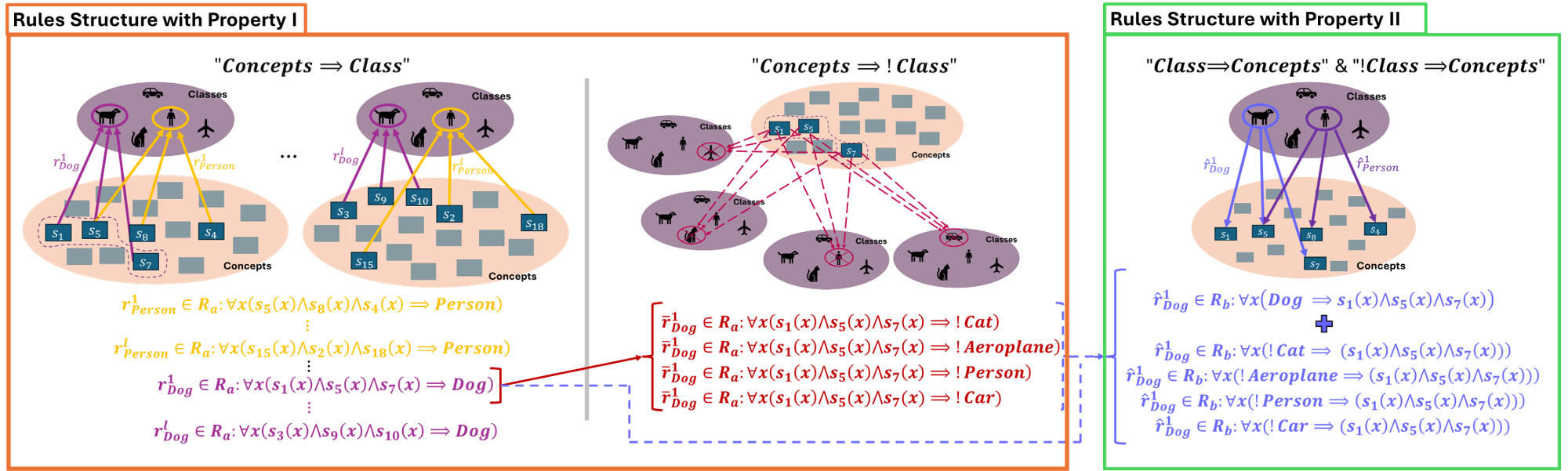}
    \caption{
Illustrates the rules structure in the knowledge base. The rules are organized into two complementary properties. \textbf{Property} I (left): Forward associations $R_a$ encodes $l$ random and distinct concept-to-class mappings per class, while their corresponding negative rules (middle; depicted for one distinct rule) encode the concept-to-¬(class) associations. \textbf{Property II}(right): Converse rules $R_b$ are derived from the forward and negative associations for each distinct class rule. Together, they form logical constraints that enforce a bidirectional structure in the rule base. 
  }
  \label{fig:RuleGen}
\end{figure*}

\begin{figure*}[t]%
  \centering
  \includegraphics[width=\textwidth]{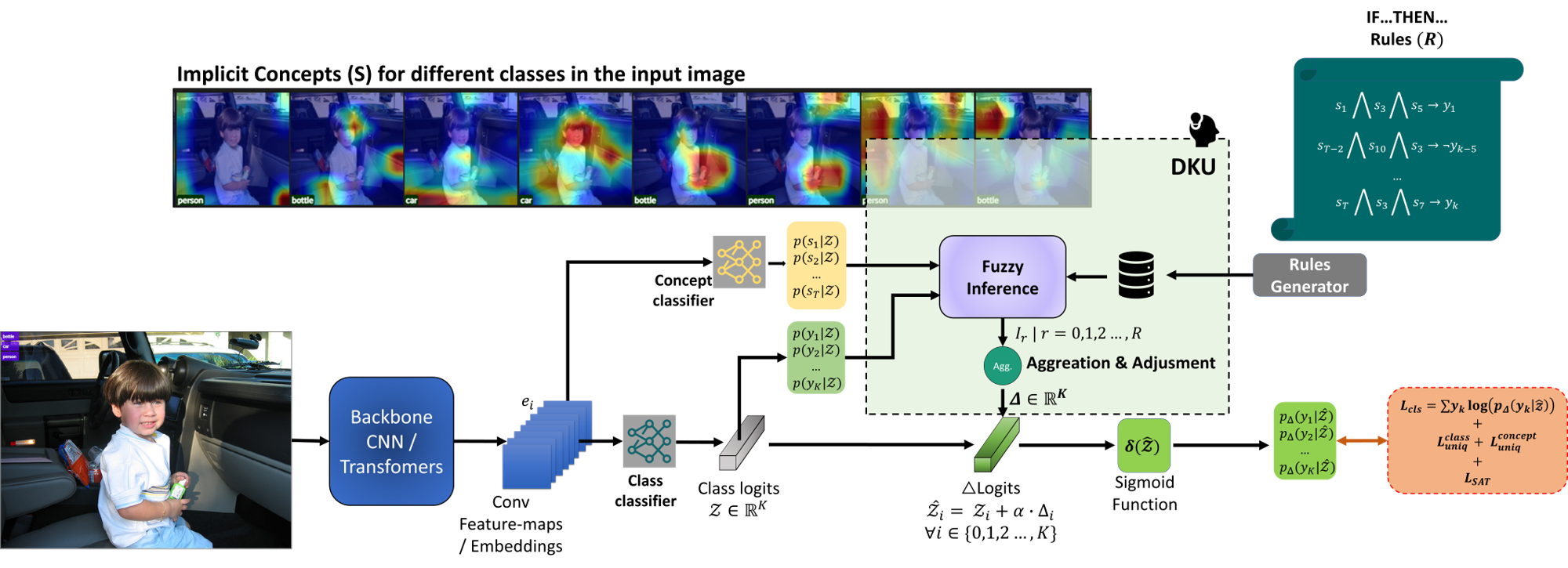}
  \caption{
  The figure presents an overview of the KLUE architecture. The DKU uses initial-class and implicit-concepts probabilities to compute the adjustment $(\Delta)$. Which enables to modulate the initial class logits ($z\rightarrow\hat{z}$) to obtain refined class probabilities $p_{\Delta}(y_k/\hat{z})$.
  The binary cross-entropy and the other auxiliary loss terms implicitly train the concepts.
  }
  \label{fig:wide-figure}
\end{figure*}

Given a dataset $\mathcal{D}=\{(x_i, y_i)\}_{i=1}^N$, $\left(x_i\in\mathcal{X}, y_i\in\mathcal{Y}\right)$, where $N$ is the total number of images and $\mathcal{Y}_i\in\{0,1\}^K$, is a multi-label ground-truth vector for the $i$-th image, where $k\in [K]$ denotes classes, our goal is to learn a robust classifier. 
This classifier should implicitly identify meaningful support concepts ($s\in[S]$) from the images and refine its predictions using symbolic logical rules defined over these concepts.
The network architecture, as shown in Fig.~\ref{fig:wide-figure}, comprises a shared backbone, two independent prediction heads, and a Differentiable Knowledge Unit (DKU). 
A shared backbone $f_{\mathrm{BB}} \colon \mathbb{R}^{H\times W\times C} \to \mathbb{R}^D$ 
maps each image to a $D$--dimensional feature vector $e_i = f_{\mathrm{BB}}(x_i) \;\in\; \mathbb{R}^{D}.$
The primary head is denoted as $h_{class}$, and is defined by the mapping function $h_{class}: \mathbb{R}^D \rightarrow \mathbb{R}^K$ computes   
the class-wise probability $p(k|h_{class}(e_i))$, which in the interest of brevity we denote it as $p(k|z)$. 
Similarly, a secondary head denoted as $h_{concept}$, defined by the mapping function $h_{concept}: \mathbb{R}^D \rightarrow \mathbb{R}^S$  computes concept-wise probability $p(s|h_{concept}(e_i))$, for brevity it is denoted as $p(s|z)$. 
The primary head is called a primary classifier, and the latter head is a support-concept classifier in our paper.
Here, $z$ denotes the pre-logit space of the respective classifiers.

The DKU takes as input the class probabilities $p(k|z)$, the probabilities of the supporting concepts $p(s|z)$, and a set of predefined rules $R_k \in \mathcal{R}$ for each class $(k)$.
The DKU then determines an adjustment vector $A \in \mathbb{R}^K$ to compute the refined class probabilities $p_{\Delta}(k|\hat{z})$.
\begin{equation}
    \label{eq::DKU}
    \Delta_k = \text{DKU}\Bigl( p(k|z), \{p(s|z) : s \in \mathcal{S}_k\}, \mathcal{R}_k \Bigr) 
\end{equation}
\begin{equation}
    \label{eq::logit scaling}
    \hat{z_k} = z_{k} + \alpha_{temp} \cdot \Delta_k 
\end{equation}
The predicted probabilities for the support concepts associated with class k are denoted as \(\{p(s|z): s \in \mathcal{S}_k\}\).
Here, $S_k\subset S$ is a subset of support concepts that are associated with class k.
\(\mathcal{R}_k\) is the set of logical rules assigned to class \(k\).
These rules consist of logical relationships between the support concepts $S_k$ and the class $k$ itself.   
The module \(\text{DKU}(\cdot)\) serves as an update mechanism that integrates the evaluation of these rules to generate $\hat{z}$, which subsequently refines the probability distribution to \(p_{\Delta}(k|\hat{z})\).
The \text{DKU} uses differentiable fuzzy-logic semantics and learnable functions to compute $\hat{z}$ during the forward pass, enabling gradient propagation and effective model optimization. The Sec.\ref{subsec::LUM} provides further details about the DKU.

\textbf{Implicit knowledge discovery:} However, the function $\text{DKU}(\cdot)$ relies on the probability estimates produced by a support-concept classifier.
An important question regarding DKU is how to determine the probabilities for the support concepts. 
Training the support concept classifier is non-obvious because the ground-truth labels for these concepts are not available. 
One of the research objectives is to develop a learning mechanism that allows the support classifier to learn from weak supervision. 
This can be thought of as discovering useful  knowledge that the DKU can integrate.
To achieve this, we create a rule-based framework that encodes the relationships between the latent support concepts and the target class label. 
Sec.~\ref{sec:RuleGen} describes further details of the rule generation. 
$r_k\in\mathcal{R}$ 
\begin{equation}
  r_{k,l}:\quad
  \bigl(\bigwedge_{a_i\in\mathcal A_{k,l}} a_i\bigr)\;\longrightarrow\; y_k
  \label{equation::horn_distinct}
\end{equation}
\noindent
where $\mathcal A_{k,l}=\{a_{k,1},\dots,a_{{k,l}}\}$, and $a_{k,l} \subset S$  is the set of concepts that make up the antecedent of $r_{k,l}$ for class $k\in K$ and $l$ is the number of rules per class.
Our idea is to utilize the logical relationship in Equation~\ref{equation::horn_distinct} between the support concepts and the classes in order to train $h_{\text{concept}}$. 
We achieve this by directly optimizing the classification loss on the refined class probabilities $p_{\Delta}$, which are updated via the $DKU$.
The intuition is that by minimizing the loss function on $p_{\Delta}$, we implicitly train $h_{\text{concept}}$.  
\begin{equation}
    \mathcal{L}_{class} = - \sum_{k=1}^{K} y_k \log p_{\Delta}(k | \hat{z})
\end{equation}
Given that $\mathcal{L}_{class}$ is minimized, the update based on the rules $\mathcal{R}_k$ must have increased the probability for the correct class $y_k$.
For the logical relationships in $\mathcal{R}_k$ as defined using Eq.(~\ref{equation::horn_distinct}), the identified support concepts must be meaningful.
Otherwise, the described implication relationship would increase $\mathcal{L}_{class}$.
Therefore, $h_{\text{concept}}$ is trained implicitly to identify concepts in the given image that are relevant for the given classes, given the rules.

\subsection{Synthetic Knowledge Base Generation}
\label{sec:RuleGen} %

To enforce the model's reasoning capabilities and promote implicit concept learning, we generated a synthetic knowledge base $\mathcal{R}$ composed of logical structure as provided in Eq.(~\ref{equation::horn_distinct}) and illustrated in Fig.~\ref{fig:RuleGen}. 
Let $\mathcal{S}=\{s_1,s_2,\dots,s_T\}$ denote set of undefined concepts and $\mathcal{Y}=\{y_1,y_2,\dots,y_K\}$ be the set of task classes respectively.
The rule base is composed of two complementary rule sets: \textbf{(a) Forward Implication $R_a$}: Represents positive and negative associations between concepts and classes. 
For each target class $y_k\in Y$, we randomly sample $l$ distinct concept combinations to generate rules of the form  "$concepts \implies class$". 
The combination length is chosen empirically.
To encourage concept overlap with other classes, individual concepts may be resampled to form distinct concept combinations for other classes, i.e., $y_j \in Y_{other\ classes} = Y \setminus \{y_k\}$, where $j \neq k$.
Figure~\ref{fig:RuleGen} (left) shows one distinct rule for "Dog" and "Person" classes, where concept "$S_5$" appears in the rule of both the classes.
For example, "$S_5$"can be any shared visual concept between the "Dog" and "Person" class. By 
having such shared property in the rule construction, we enable the model to learn the shared attributes between the classes. 
However, the assignment of shared concepts is randomized in our rule generation step.
Correspondingly,  negative associations are derived under the assumption that concept combination supporting a target class $y_k$ imply conflicting relationships with non-target classes $Y_{non-target} = Y \setminus\{y_k\}$. This enforces the constraint that a unique set of concepts implies only a single class.
Figure \ref{fig:RuleGen} (middle)  illustrates such negative rules derived from one distinct positive rule. 
\textbf{(b) Inverse implications $R_b$}: These rules encode class-to-concepts (converse rule) relation derived from the distinct positive and negative forward associations (i.e., from $R_a$). 
Each class implies the union of concepts that imply the class in any of its associated implication rules.
Together, $R_a$ and $R_b$  forms the synthetic rule base $\mathcal{R}=\{ R_a, R_b\}$, encoding bi-directionality. 
For the concrete algorithm to construct the knowledge base, cf. the supplementary material.

\subsection{Differentiable Knowledge Unit}
\label{subsec::LUM}
The purpose of the differentiable knowledge unit $DKU$ is to estimate the adjustment $\Delta$ to attenuate the original logits $z$ of the primary classifier, such that it refines the probability estimate $p_\Delta(k|\hat{z})$.
As Eq.(~\ref{eq::DKU}) describes, it takes $p(k|z)$, a set of rules $\mathcal{R}_k$ associated with class $k$, and the concept probabilities $p(s|z)$ as input. 
The DKU integrates symbolic knowledge, expressed via fuzzy logic,  into deep learning models. 
The DKU is structured around the following fundamental principles: rule-based knowledge representation, probabilistic grounding, and fuzzy inference.
Knowledge is captured using a structured rule base $\mathcal{R}$, where each rule "$R_a\subset\mathcal{R}$" following the canonical Eq.(~\ref{equation::horn_distinct}), and their inverse rules "$R_b\subset\mathcal{R}$" are grounded using fuzzy relaxation \cite{hajek2001metamathematics,donadello2017logictensornetworkssemantic, AnalyzingDFL} to enable smooth differentiation.
Antecedents represent conjunctions of concepts drawn from a support concept set  
$S$.Whereas,
Consequent corresponds to the class from the set 
$K$ in $R_a$. However, the rules are the other way around in $R_b$.
Each support concept $s_i$ can occur in multiple rules. 
If a support concept occurs in multiple rules, and these rules are associated with different classes. In that case, the optimization enforces the support concept, requiring that the support of each class be unequal.
Probabilistic grounding in the DKU associates truth values of concepts and classes directly with corresponding neural network classifier outputs. 
Each concept $s_i$ has an associated truth value $p(s_i|z)$, estimated from a dedicated concept classifier.
Similarly, category truth values are grounded using the primary classifier's estimate  $p(k|z)$.

\textbf{Evaluating the Concept Combination:} To obtain the membership of the set of concepts $a_i\subset \mathcal{A}_{k,l}$ in Eq.(~\ref{equation::horn_distinct}) belonging to individual rules, we explore two different fuzzy logic connectors. 
The first is the parameterized activation connector that we propose:
\begin{equation}
    \label{eq:parametericFuzzyAcitvation}
    A(x,y) = \sigma(\alpha x + \beta y + \gamma x\cdot y + \delta)
\end{equation}
where the parameters $\alpha, \beta, \gamma$ and $\delta$ are optimized during network training, and $\sigma(\cdot)$ is a sigmoid function. 
Whereas, the second connector is the Yager t-norm \cite{YAGER1980235}.
The parameterized activation connector is a learnable operator that approximates a fuzzy operator, enabling it to learn a smooth and differentiable function during optimization.

\textbf{Evaluating the implication:} 
Given the antecedent and consequent truth-ness for each rule $r_{k,l} \in R$, the rule satisfaction is computed using the Implication functions. 
We explore two different implication operations.
On the one hand, Reichenbach Implication \cite{Nagel_1941} and on the other hand, the Sigmoidal Reichenbach Implication proposed in \cite{AnalyzingDFL}.
The main difference between these two functions lies in their smoothness and differentiability.

\textbf{Aggregation:} As mentioned in Sec.~\ref{sec:RuleGen}, each class has $M$ and $N$ positive and negative rules. 
Therefore, the truthness of all associated rules should be aggregated in a meaningful way to obtain the adjustment factor $\Delta_k$ for each class $K$. To facilitate rule truthness aggregation, contrary to Softmin(\cite{manhaeve2018deepproblog})  we use the following function softmax-weighted average aggregator Eq.(~\ref{eq:softmaxWAagg}):
\begin{equation}
\label{eq:softmaxWAagg}
\begin{aligned}
\text{Softmax‑WA: }Agg\bigl(I^{\pm}_{k}\bigr)
&=
\sum_{i}
\frac{\exp\bigl(\tau\,t_{i}^{\pm}\bigr)}
     {\displaystyle\sum_{j}\exp\bigl(\tau\,t_{j}^{\pm}\bigr)}
\;t_{i}^{\pm}\,. 
\end{aligned}
\end{equation}

where $t_i$ and $t_j$ represent the individual truth values of rules for a given class $k$.
For each class  $y_k$, the category-specific adjustment $\Delta_k$ is computed as the difference between aggregated positive and negative truth values.
\begin{equation}
    \label{eq:adjustment}
    \Delta_k = Agg(I^{+}_{k}) - Agg(I^{-}_{k})
\end{equation}

Finally, predictions from the primary classifier are refined by adjusting its logits  with the calculated adjustment terms across all categories: $\hat{z} = z + \alpha_{temp} \cdot \Delta$.
A sigmoid layer $\sigma(\cdot)$ transforms the updated logits $\hat{z}$ to the final probabilities: $p_\Delta(y_k|\hat{z}) = \sigma(\hat{z})$.
Through these principles, the DKU effectively leverages the implicit support concepts learnt via the defined rule construct to enhance performance in the classification task. 

\subsection{Enforcing the Rule Base Properties}
\label{subsec::unique}
For our rules, the support concepts must represent unique categories. 
I.e., they must be different from one another and also represent different categories from the classes. 
Otherwise, the implicit optimization of the support categories could fall into a local minimum, where the respective class is classified again, since a class is completely causally related to itself.
We tackle this problem by designing a uniqueness loss. 
Our classifier heads for the support concepts and our classes are a single fully connected layer that takes the shared latent space as input. %
Since the weights all operate on the same input, we can identify the similarities of the classes/concepts expressed by these classifiers directly by comparing their weights. 
We use cosine similarity to compare the weights and aim to minimize it. 
In mathematical terms, if we let 
$\tilde{W_K} \in \mathbb{R}^{K \times d}$
denote the matrix whose rows are the normalized weight vectors of $h_{class}$ $\tilde{w}_k = \frac{w_k}{\|w_k\|_2}$, and $\tilde{W_S} \in \mathbb{R}^{S \times d}$ vice versa is the matrix that represents the normalized weight of $h_{concept}$.
Then we can compute the two loss terms based on cosine similarity.
\begin{equation}
    \mathcal{L}^{concept}_{\text{uniq}} = \|\tilde{W}_S\tilde{W}_S^\top - I\|_F^2
\end{equation}
which enforces orthogonality between the weights of the concept classifiers themselves ($I$ identity matrix) and
\begin{equation}
    \mathcal{L}^{class}_{\text{uniq}} = \|\tilde{W}_S\tilde{W}^\top_K \|_F^2
\end{equation}
which enforces orthogonality between the concept classifiers and the class classifiers.

\paragraph{Bidirectional relationship:}
We enforce a bidirectional relationship between the classes and their support concepts.
This property of the rule base ensures that each class implies the conjunction of its support concepts.
Formally, for each class \(k\) in the give training sample, converse rules $R_b \subset \mathcal{R}$ of the nature"
\(\bar{r}_k:  y_k \!\rightarrow\!\bigl(\bigwedge_{a_i\in\mathcal A_{k,l}} a_i\bigr)\)" is considered.  
We evaluate the truth degree of these rules using the previously defined fuzzy logic operators and SAT-Aggregator proposed in \cite{LTN2022} to obtain the Satisfiability $SAT_{\bar{r}}$ degree of valid rules. 

\begin{equation}
\label{eq:bid-loss}
\mathcal{L}_{\text{SAT}}
=
1- SAT_{\bar{r}}
\end{equation}

\subsection{KLUE Assembling the Bricks}
\label{sec:AssemblingBricks}
We name our method \textbf{KLUE} (\textbf{K}nowledge and \textbf{L}ogic \textbf{U}pdate for \textbf{E}nhanced Recognition). 
To integrate KLUE into a given architecture, one can follow the steps described in the previous sections:
First, we need to generate a set of rules for each given class as described in Sec.~\ref{sec:RuleGen}. 
Second, we integrate the DKU (see Sec.~\ref{subsec::LUM}) to compute the $\Delta$ to update the initial classifier probabilities. 
Depending on the logic functions, we define two versions of KLUE, which have different implementations of the conjunction, implication, and aggregation functions: 
KLUE(v1): conjunction Eq.(~\ref{eq:parametericFuzzyAcitvation}),  Reichenbach implication and aggregation Eq.(~\ref{eq:softmaxWAagg}) and
KLUE(v2): conjunction yager tnorm and sigmoidal Reichenbach implication and aggregation Eq.(~\ref{eq:softmaxWAagg}).
The training of a network with KLUE integrated utilizes the following loss function:
\begin{equation}
    \mathcal{L}_{class} + \lambda_{class} \mathcal{L}_{\text{uniq}}^{class} + \lambda_{concept} \mathcal{L}_{\text{uniq}}^{concept} +
    \lambda_{SAT} \mathcal{L}_{\text{SAT}}
\end{equation}
Where $\lambda_{class}$, $\lambda_{concept}$, and $\lambda_{SAT}$ influence the weight that the rules' properties have in the optimization process.
As described in Sec.~\ref{subsec::unique}, $\mathcal{L}_{class}$ implicitly trains the concept classification heads and thereby enables the discovery of feasible knowledge to integrate with the rules.

\section{Experiments}
\label{sec:Experiments}
Our experiments are designed to test the robustness that is introduced by our KLUE approach. 
In order to do so, we provide the classification results on three different datasets:
Although PASCAL VOC 2012~\cite{Pascal} and COCO~\cite{COCO} are primarily object-detection benchmarks with bounding boxes, in our experiments, we repurpose them for multi-label classification by treating each image’s annotation as a binary presence/absence vector.
Additionally, we provide ablation studies on the medical ChestMNIST dataset, which is part of the MNISTv2~\cite{yang2023medmnist} collection. 
It also represents a multi-label classification task with 14 binary class decisions. 
These datasets represent two domains of application: natural and medical gray value images. 
The backbone networks we use are the WideResNet-101~\cite{zagoruyko2016wide} (WRN-101) and the Swin-V2-Tiny~\cite{liu2022swin}(SWIN) architecture. 
The vanilla version of these models with standard loss is considered a baseline.
We integrate KLUE into the decision layer of these networks and tag these models as KLUE variants. 
The results are compared using the mean average precision (mAP) with thresholds between 0.5 and 0.95, and the area under the receiver operating characteristic curve (AUC) on the validation set of the respective datasets. 
All our experiments are run with 3 seeds to ensure reliable results. 
If the Optuna framework \cite{akiba2019optuna} is not used for the hyperparameter finetuning as in Sec.~\ref{ssec:KLUEv2v3}, then we choose the learning rate (LR), weight decay (WD), and learning rate step size (SZ) to be:
Wide ResNet-101 LR:$1e-4$ SZ:$7$ WD:$1e-5$,
Swin LR:$5e-5$ WD:$1e-5$.
Unless otherwise specified, we employ 100 concept classifiers, with $5$ positive and $95$ negative rules for PASCAL VOC. 
For COCO, with 80 classes, this yields 395 negative rules. As explained in Sec.~\ref{sec:RuleGen}, each positive rule for a class generates negative rules for all other classes.
The rule system is generated as described in Sec.~\ref{sec:RuleGen}.
We set the temperature parameter $\alpha_{temp}$ to be $5.0$ the  orthogonality losses $\lambda_{concepts}:0.1$, $\lambda_{class}:0.01$, and $\lambda_{SAT}:1.0$.

\begin{table}[t]
    \centering
    \caption{Average mAP and AUC (3 seeds) on the validation set of COCO and PASCAL VOC 2012. \textbf{$\mathsf{E}$}: use of external rules $R_{ext}$.}
    \label{tab:combined_all}
    \resizebox{\linewidth}{!}{%
    \begin{tabular}{@{}l l cc cccc@{}}
      \toprule
      \textbf{Enc.} & \textbf{Model}
        & \textbf{$\mathcal{L}_{\text{SAT}}$}
        & \textbf{\textbf{$\mathsf{E}$}}
        & \multicolumn{2}{c}{\textbf{COCO}}
        & \multicolumn{2}{c}{\textbf{PASCAL VOC}} \\
      \cmidrule(lr){5-6}\cmidrule(lr){7-8}
        &  &  &  & \textbf{mAP} & \textbf{AUC} & \textbf{mAP} & \textbf{AUC} \\
      \midrule

      \multirow{7}{*}{\rotatebox[origin=c]{90}{\textbf{WRN-101}}}
        & Baseline           & \xmark & \xmark & 71.97 & 96.83 & 87.73 & 97.76 \\
        & LTN (v1)           & \xmark & \cmark & 41.08 & 92.46 & 88.74 & 97.97 \\
        & LTN (v2)           & \xmark & \cmark & 40.84 & 92.40 & 87.70 & 97.69 \\
        & KLUE (v1)          & \xmark & \xmark & 74.06 & 97.08 & 89.24 & 98.02 \\
        & KLUE (v2)          & \xmark & \xmark & 72.96 & 97.05 & 88.07 & 97.75 \\
        & KLUE (v1)          & \cmark & \xmark & 80.79 & 97.88 & \textbf{91.32} & \textbf{98.37} \\
        & KLUE (v2)        & \cmark & \xmark & \textbf{80.80} & \textbf{97.91} & 90.22 & 98.10 \\
      \midrule

      \multirow{7}{*}{\rotatebox[origin=c]{90}{\textbf{SWIN}}}
        & Baseline           & \xmark & \xmark & 82.24 & 98.29 & 91.32 & 98.60 \\
        & LTN (v1)           & \xmark & \cmark & 80.47 & 98.03 & 91.00 & 98.65 \\
        & LTN (v2)           & \xmark & \cmark & 79.46 & 97.64 & 91.81 & \textbf{98.75} \\
        & KLUE (v1)          & \xmark & \xmark & \textbf{82.78} & \textbf{98.30} & 92.30 & 98.65 \\
        & KLUE (v2)          & \xmark & \xmark & 81.93 & 98.13 & 92.10 & 98.58 \\
        & KLUE (v1)        & \cmark & \xmark & 82.71 & 98.23 & \textbf{92.38} & 98.58 \\
        & KLUE (v2)        & \cmark & \xmark & 82.22 & 98.13 & 92.02 & 98.51 \\
      \bottomrule
    \end{tabular}}
    {\footnotesize \raggedright \textit{Note:} v1/v2 denote different fuzzy operator instantiations (see Sec.~\ref{sec:AssemblingBricks}) \par}
\end{table}

Given that in our scenario no external rules are provided, finding a direct comparison method is difficult, as, to our knowledge, no other framework can automatically derive rules.  
Therefore, to position our proposed approach's performance with the NeSy approach, we consider the LTN variants of the above models, which are optimized with standard loss and $SAT_{\text{LOSS}}\cite{LTN2022}$ on external mined rules $\mathcal{R}_{ext}$. These rules constitute statistical correlations expressed as associations between classes \cite{Apriori}.

\subsection{Results on PASCAL VOC and COCO}
\label{sec:main_results:PASCALCOCO}
Table~\ref{tab:combined_all} reports the mean average precision (mAP) and area under the ROC curve (AUC), averaged over three random seeds, for all models on PASCAL VOC 2012, MS COCO.
Across both datasets, our KLUE architecture consistently outperforms the WRN-101 baseline and the LTN variant. 
The effect is especially strong for the WRN-101, where the best KLUE variant improves the mAP by $12.27\%$ and the AUC by $1.11\%$ relative to the baseline on COCO.
Whereas, on PACAL VOC 2012, it outperforms by $4.09\%$ mAP and $0.2\%$ AUC.
However, for the SWIN-based KLUE models, the improvements are not as strong as those of the WRN-101 KLUE versions. Nonetheless, the mAP is better than the baseline and LTN variants. 
For qualitative results, cf. supplementary.  

\subsection{Domain Generalization}
\label{sec:domain_generalization}

\begin{table}[t]
    \centering
    \caption{Domain generalization performance: Mean Average Precision (mAP) and AUC of models trained on PASCAL VOC and evaluated on the COCO validation set. 
    WRN-101: WideResNet-101; Swin: Swin-V2-Tiny. \textbf{$\mathsf{E}$} use of external rules.}
    \label{tab:generalization}
    \renewcommand{\arraystretch}{1.1}
    \resizebox{0.9\linewidth}{!}{%
    \begin{tabular}{@{}l cc cc cc@{}}
      \toprule
      \textbf{Model} 
        & \textbf{$\mathcal{L}_{\text{SAT}}$} 
        & \textbf{$\mathsf{E}$} 
        & \multicolumn{2}{c}{\textbf{WRN-101}} 
        & \multicolumn{2}{c}{\textbf{SWIN}} \\
      \cmidrule(lr){4-5} \cmidrule(lr){6-7}
        &  &  & \textbf{mAP} & \textbf{AUC} & \textbf{mAP} & \textbf{AUC} \\
      \midrule
      Baseline & \xmark & \xmark & 64.26 & 92.44 & \textbf{72.51} & \textbf{94.95} \\
      LTN      & \xmark & \cmark & 63.54 & 92.24 & 71.63 & 94.82 \\
      KLUE     & \xmark & \xmark & 68.30 & 93.55 & 72.44 & 94.41 \\
      KLUE   & \cmark & \xmark & \textbf{74.44} & \textbf{94.32} & 72.40 & 94.65 \\
      \bottomrule
    \end{tabular}}
\end{table}

One of the primary motivations for introducing knowledge into neural networks is to enhance their ability to generalize.
We use the generalization across unseen data domains to test this feature of KLUE.
To assess how well our models transfer beyond the source domain, we train on PASCAL VOC 2012 and evaluate on the COCO validation set, utilizing the overlapping classes.  
Table~\ref{tab:generalization} reports the mean average precision (mAP) and ROC AUC obtained on this unseen domain.
The WRN-101 baseline achieves an mAP of 64.26\% and an AUC of 92.44\%.
Our best KLUE version of WRN-101 boosts mAP by $15.84\%$ and AUC by $2.03\%$ relative to the baseline version.
Transformer-based architectures generally have better generalization performance, as reflected in Tab.~\ref{tab:generalization}.
The SWIN baseline is slightly better than the other variants. However, the LTN-enhanced SWIN version shows a drop in mAP. In contrast, the KLUE-based SWIN version matches the baseline's performance. 
On the one hand, Tab.~\ref{tab:generalization} shows that the LTN variant performs worse than other model variants.
This observation provides empirical evidence that the externally mined rules do not transfer well to unseen domains, validating our argument that such rules often constitute untargeted knowledge. 
On the other hand, the model variants with the KLUE bottleneck demonstrate either a significant performance gain over the baseline( WRN-101) or comparable performance (SWIN), indicating that learning implicit concepts under a logical inductive bias helps the model capture targeted and semantically meaningful rules that generalize better.

\subsection{Robustness towards Corner Case Samples}
\label{sec:corner_cases}

\begin{table}[t]
    \centering
    \caption{Performance of WideResNet-101 (WRN-101) and KLUE-WideResNet-101 (KLUE-WRN-101)
    on the PASCAL VOC full validation set and 90th percentile high-entropy samples. \textbf{$\mathsf{E}$} use of external rules.}
    \label{tab:hard-samples}
    \renewcommand{\arraystretch}{1.1}
    \resizebox{1\linewidth}{!}{%
    \begin{tabular}{@{}l cc cc cc@{}}
      \toprule
      \textbf{Model} 
        & \textbf{$\mathcal{L}_{\text{SAT}}$} 
        & \textbf{$\mathsf{E}$} 
        & \multicolumn{2}{c}{\textbf{Full Validation Set}} 
        & \multicolumn{2}{c}{\textbf{Hard Samples}} \\
      \cmidrule(lr){4-5} \cmidrule(lr){6-7}
        &  &  & \textbf{mAP} & \textbf{AUC} & \textbf{mAP} & \textbf{AUC} \\
      \midrule
      Baseline & \xmark & \xmark & 87.84 & 97.76 & 79.75 & 96.46 \\
      LTN      & \xmark & \cmark & 88.75 & 97.96 & 85.67 & 96.15 \\
      KLUE     & \xmark & \xmark & 89.20 & 98.02 & 87.03 & 97.11 \\
      KLUE   & \cmark & \xmark & \textbf{91.32} & \textbf{98.40} & \textbf{89.96} & \textbf{98.16} \\
      \bottomrule
    \end{tabular}}
\end{table}

Apart from improvements within a data domain and for domain generalization, the integration of knowledge is expected to improve robustness to corner-case data. 
To measure corner‐case performance, we split the PASCAL VOC validation set into (i) the full validation set, and (ii) the top 10\% most uncertain images under the baseline model (hard samples).
The latter are determined by the baseline model's entropy, which measures the network's uncertainty. 
Table~\ref{tab:hard-samples} compares the baseline WRN-101 and the KLUE variant of WRN-101.
For the full validation set, KLUE improves the mAP by around 4\% compared to the baseline.
For the hard sample set, the baseline mAP drops by 9.2\% to 79.75\%, whereas the KLUE model performs very similarly on the hard samples and the full set, at 91.32\% vs. 89.96\%.
Hence, KLUE loses very little performance on the hard samples and is even roughly on par with the baseline's performance on the full validation set. 
Here, the bijective property of the rule base enforced by $\mathcal{L}_{\text{SAT}}$ is especially relevant.
Without this property, even KLUE performance drops by 2.4\% compared to the full validation set, but it is still better than LTN. 
The external rules integrated into the LTN also enhance robustness. 
Our more targeted, automatically discovered rules, however, perform better. 
These results demonstrate that the knowledge self-discovered by KLUE not only improves in-domain performance but also generalizes to novel distributions, thereby increasing robustness to corner cases.

\subsection{KLUE v1 and v2}
\label{ssec:KLUEv2v3}

\begin{table}[t]
    \centering
    \caption{
    Comparison of KLUE (v1) and KLUE (v2) w.r.t. mean Average Precision (mAP) 
    and AUC on PASCAL VOC 2012 and ChestMNIST. 
    All models use a WRN-101. 
    \textbf{$\mathsf{E}$} use of external rules.}
    \label{tab:optuna_mAP_on_pascalvoc2012ChestMNIST}
    \renewcommand{\arraystretch}{1.1}
    \resizebox{0.95\linewidth}{!}{%
    \begin{tabular}{@{}l cc cc cc@{}}
      \toprule
      \textbf{Model}
        & \textbf{$\mathcal{L}_{\text{SAT}}$}
        & \textbf{$\mathsf{E}$}
        & \multicolumn{2}{c}{\textbf{PASCAL VOC 2012}}
        & \multicolumn{2}{c}{\textbf{ChestMNIST}} \\
      \cmidrule(lr){4-5} \cmidrule(lr){6-7}
        & & & \textbf{mAP} & \textbf{AUC} & \textbf{mAP} & \textbf{AUC} \\
      \midrule
      WRN-101       & \xmark & \xmark & 89.65 & 98.14 & 23.0 & 82.3 \\
      KLUE (v1)     & \xmark & \xmark & 92.34 & \textbf{98.65} & 23.7 & 83.0 \\
      KLUE (v2)     & \xmark & \xmark & 90.27 & 98.21 & 23.8 & \textbf{83.1} \\
      KLUE (v1)   & \cmark & \xmark & \textbf{92.63} & 98.62 & 23.6 & 82.5 \\
      KLUE (v2)   & \cmark & \xmark & 91.50 & 98.47 & \textbf{24.2} & 83.0 \\
      \bottomrule
    \end{tabular}}
\end{table}

As Sec.~\ref{sec:AssemblingBricks} describes, we have different instantiations of KLUE. 
They differ in how the fuzzy logic operators are instantiated.
To compare their performance fairly with each other, we employed the Optuna framework to speed up the search over the hyperparameters: LR, SZ, WD, $\alpha_{\mathrm{temp}}$, orthogonality loss weights $\lambda_{concepts}$ and $\lambda_{class}$, and $\lambda_{SAT}$ for WRN-101 with KLUE. 
Table~\ref{tab:optuna_mAP_on_pascalvoc2012ChestMNIST} shows the mAP achieved after tuning for the PASCAL VOC 2012 dataset and the ChestMNIST dataset. 
Both KLUE versions outperform the baseline, even with substantial effort invested in hyperparameter optimization.
Since both versions of KLUE (v1) and (v2) yield an improvement, this suggests a certain robustness of the framework towards the implementation of fuzzy logic. 
KLUE(v1) performs better for the PASCAL VOC dataset, while KLUE(v2) performs better for the ChestMNIST data. 
Generally speaking, there is no clear winner; however, taking the values from Tab.\ref{tab:combined_all} into account, one can conjecture that (v1) performs better on natural images and
v2 has an advantage in medical images.

\subsection{Hyperparameters of the Rulebase}
\label{sec:ablation_num_concepts_pascal}

\begin{table}[t]
    \centering
    \caption{The mAP values on the validation set of PASCAL VOC for different concept numbers in KLUE. Backbone WRN-101.}
    \label{tab:transposed_sorted}
    \resizebox{\linewidth}{!}{%
    \begin{tabular}{l c cccccccc}
        \toprule
        & \textbf{$\mathcal{L}_{\text{SAT}}$} 
        & \#1 & \#2 & \#3 & \#4 & \#5 & \#6 & \#7 & \#8 \\
        \midrule
        num\_concepts          & \xmark & 10 & 20 & 100 & 120 & 130 & 140 & 150 & 160 \\
        KLUE(v1)         & \xmark & 92.17 & \textbf{92.22} & 92.18 & 92.08 & 92.15 & 92.17 & 92.17 & 92.10 \\
        KLUE(v1)       & \cmark & 92.44 & 92.32 & \textbf{92.51} & 92.30 & 92.39 & 92.39 & 92.17 & 92.38 \\
        \bottomrule
    \end{tabular}}
\end{table}

\begin{figure}[t]
    \centering
    \includegraphics[width=\linewidth]{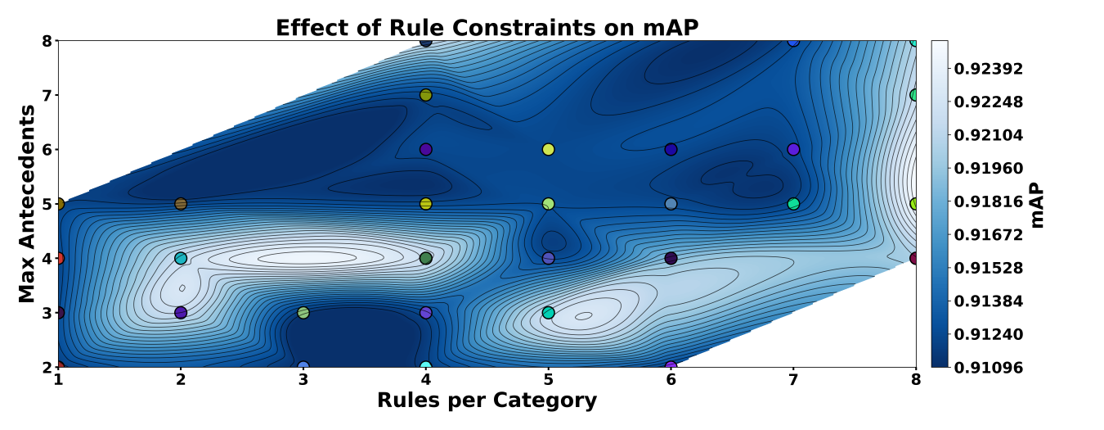} %
    \caption{Functional relationship between the rules per category and the maximum number of concepts in the antecedence}
    \label{fig:QualitativeExamples}
\end{figure}
The number of concepts utilized to create the rules is an interesting parameter, as it represents the "parameters" of the Differentiable Knowledge Unit (DKU). 
We therefore present an ablation study by varying the number of learned concepts in our DKU from 10 to 160.
Table~\ref{tab:transposed_sorted} reports the validation mAP for each setting on the PASCAL VOC dataset. 
One can observe that the overall difference in mAP over the number of concepts is rather small, both for KLUE (v1) and (v2).
This suggests a certain robustness towards the hyperparameter "number of concepts". 
But it also suggests a certain redundancy in the meaning of the concepts discovered by KLUE.
Future work is to investigate loss functions to make the concepts more diverse. 
Figure~\ref{fig:QualitativeExamples} shows how the number of concepts in the antecedence interacts with the number of positive rules per category with respect to the mAP. 
One can see that these hyperparameters of the rule base are quite robust w.r.t variations, as there are little absolute differences between the tested combinations (range 92.4\%-91.2\%). This indicates KLUE's robustness.

\subsection{Prevention of Overfitting}
\label{sec:overfitting}
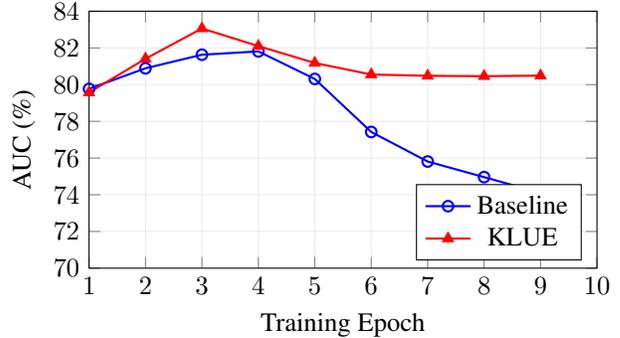
\begin{figure}[t]
  \centering
  \begin{tikzpicture}
    \begin{axis}[
      width=\columnwidth,
      height=5cm,                 %
      xlabel={Training Epoch},
      ylabel={AUC (\%)},
      xmin=1,   xmax=10,
      ymin=70,   ymax=84,
      xtick={1,2,3,4,5,6,7,8,9,10},
      ytick={70,72,74,76,78,80,82,84},
      grid=both,
      major grid style={opacity=0.3},
      minor grid style={opacity=0.1},
      legend pos=south east,
    ]

      \addplot[
        thick,
        mark=o,
        color=blue,
      ] coordinates {
        (1,79.77)%
        (2,80.89)%
        (3,81.63)%
        (4,81.81)%
        (5,80.31) %
        (6,77.42) %
        (7,75.81) %
        (8,74.96) %
        (9,74.1) %
      };
      \addlegendentry{Baseline}

      \addplot[
        thick,
        mark=triangle*,
        color=red,
      ] coordinates {
        (1,79.57)%
        (2,81.41)%
        (3,83.06)%
        (4,82.1)%
        (5,81.18)%
        (6,80.55)%
        (7,80.48)%
        (8,80.46)%
        (9,80.49)%
      };
      \addlegendentry{KLUE}

    \end{axis}
  \end{tikzpicture}
  \caption{AUC over training steps for Baseline vs.\ KLUE for the ChestMNIST validation set. Averaged over 3 seeds.
  }
  \label{fig:map-valiadtion-curve}
\end{figure}

Figure ~\ref{fig:map-valiadtion-curve} shows the validation AUC of the baseline and KLUE on the ChestMNIST over the epochs. 
One can observe that the baseline after epoch 5 declines quickly, whereas KLUE remains much more stable. 
This effect is especially useful when no validation set is available for model selection. 
It is more likely to choose a bad baseline model than a good one for KLUE. 
Such a case is the domain generalization for which we have shown superior performance.
This further shows that integrating the self-discovered knowledge makes the classification more robust.

\section{Discussion}
Integrating knowledge into deep neural networks is challenging, not only because of the methodology for integrating rules and logic, but also because the knowledge to be integrated must be relevant. 
Especially for perception tasks, such knowledge is seldom available. 
This work introduces KLUE, a framework that enables networks to discover knowledge that optimally integrates within a given set of rules.
This work has demonstrated KLUE's ability to enhance both in-domain and out-of-domain generalization, as well as the increased robustness it provides against corner-case images and overfitting.
Since KLUE, unlike other neuro-symbolic AI systems, does not require external knowledge, it can be applied to any classification problem.
KLUE can also be thought of as an architecture. 
KLUE can generate a logic update for feature maps.

\section*{Acknowledgement}
This work was funded by the Horizon Europe programme of the European Union, under grant agreement 
101146542 
(project SYNERGIES). Views and opinions expressed here are however those of the author(s) 
only and do not necessarily reflect those of the European Union or CINEA. Neither the European Union nor 
the granting authority can be held responsible for them.
{
    \small
    \bibliographystyle{ieeenat_fullname}
    \bibliography{main}
}

\clearpage
\setcounter{page}{1}
\maketitlesupplementary

\section{Synthetic Rule Base Generation}
\begin{algorithm2e}[h] 
\caption{Constructing Rule Base $\mathcal{R}$}
\label{alg:rule_generation}

\DontPrintSemicolon
\SetKwFunction{FRandInt}{RandomInt}
\SetKwFunction{FSample}{Sample}
\SetKwFunction{FChoice}{Choice}
\SetKwFunction{FShuffle}{Shuffle}
\SetKwFunction{FMain}{Main}
\SetKwProg{Proc}{Procedure}{}{}

\SetKwInOut{Input}{Input}
\Input{
    $T$: Number of implicit concepts, $K$: Target classes,
    $S$: Concepts $\{s_1, \dots, s_T\}$,   
    $Y$: Classes $\{y_1, \dots, y_K\}$,
    $l$: Rules per category (\ie $concepts \Rightarrow class$), and
    $q_{min}, q_{max}$: min/max concept combination set size
}

\Proc{\FMain{$S, Y, l, q_{min}, q_{max}$}}{
    $R_a \gets \emptyset$, $R_b \gets \emptyset$;\\
    $O \gets \{s \mapsto 0 \mid \forall s \in S\}$;\\
    
    \tcp{Phase 1: Core phase}
    \ForAll{$y_i \in Y$}{
        \For{$j \gets 1$ \KwTo $l$}{
            $q \gets \FRandInt(q_{min}, q_{max})$;\\
            $A\gets \FSample(S, q)$; \tcp*{Sample concepts}
            \lForEach{$s \in A$}{$O[s] \gets O[s] + 1$;}
            
            $R_a \gets R_a \cup \{A \implies \{y_i\}\}$;
            $R_b \gets R_b \cup \{\{y_i\} \implies A\}$;
            
            \uIf{$P_{neg} > 0$}{
                $Y_{other} \gets Y \setminus \{y_i\}$;\\
                $N_{neg} \gets \lfloor P_{neg} \cdot |Y_{other}| \rfloor$;\\
                $\hat{Y}_{neg} \gets \FSample(Y_{other}, N_{neg})$;\\
                \ForAll{$y_{neg} \in \hat{Y}_{neg}$}
                {
                    $R_a \gets R_a \cup \{A\implies \{\neg y_{neg}\}\}$;
                    $R_b \gets R_b \cup \{\{\neg y_{neg}\} \implies A\}$;
                }
            }
        }
    }
    
    \tcp{Phase 2: concept-coverage phase}
    \(S_{unused} \gets \{s \in S \mid O[s] = 0\};\)\\
    \ForAll{$s^* \in S_{unused}$}{
        $q \gets \FRandInt(q_{min},q_{max})$;\\
        $A' \gets \FSample(S \setminus \{s^*\}, \max(0, q-1))$;\\
        $A \gets A' \cup \{s^*\}$;\\
        $y^* \gets \FChoice(Y)$; \\
        $R_a \gets R_a \cup \{A \implies \{y^*\}\}$; \\
        $R_b \gets R_b \cup \{\{y^*\} \implies A\}$; \\
        \lForEach{$s \in A$}{$O[s] \gets O[s] + 1$}
    }
    
    $\mathcal{R} \gets R_a \cup R_b$\\
    \Return $\FShuffle(\mathcal{R})$
}
\end{algorithm2e}

To provide a logical inductive bias for the proposed methodology, we synthesize a knowledge base, denoted as $\mathcal{R}$. 
This rule base guides the network to learn implicit support concepts by enforcing meaningful relationships between abstract concepts and downstream classes. 
The synthesis procedure, outlined in Algorithm \ref{alg:rule_generation}, is divided into two phases: a \textbf{Core Phase} to establish foundational class-concept links, and a \textbf{Concept-Coverage Phase} to ensure semantic grounding of all available concepts.

Let $S = \{s_1, s_2, \dots, s_T\}$ denote the set of abstract concepts and $Y = \{y_1, y_2, \dots, y_K\}$ denote the set of target classes. 
The final rule base is the union of compositional rules and their converses, $\mathcal{R} \gets R_a \cup R_b$.

\subsection{Phase 1: Core Rule Generation}
In the initial phase, we generate a minimum of $l$ distinct positive rules for each class $y_k \in Y$. For the $l$-th rule of class $y_k$, we determine the antecedent size $q$ by sampling uniformly from the integer interval $[q_{min}, q_{max}]$. In our experiment setup, we fix $[q_{min}=2, q_{max}=4]$. We then sample a support set of concepts $\mathcal{A}_{k,l} \subset S$ such that $|\mathcal{A}_{k,l}| = q$, strictly without replacement.

This support set $\mathcal{A}_{k,l}$ forms the basis for \textbf{compositional rules} ($R_a$), as illustrated in Figure \ref{fig:RuleGen} (left and middle panels).
First, we define the \textit{positive implication} that maps the conjunction of selected concepts to the specific category:
\begin{equation}
    \left( \bigwedge_{s \in \mathcal{A}_{k,l}} s \right) \implies y_k.
    \label{eq:positive_implication}
\end{equation}
To enforce mutual exclusivity among classes—ensuring a specific concept combination maps uniquely to one category—we explicitly construct \textit{negative associations} against all other classes $y_m \in Y \setminus \{y_k\}$:
\begin{equation}
    \left( \bigwedge_{s \in \mathcal{A}_{k,l}} s \right) \implies \neg y_m, \quad \forall m \neq k.
    \label{eq:negative_implication}
\end{equation}
Furthermore, to impose a bidirectional logical structure, we generate \textbf{converse rules} ($R_b$). 
For every positive and negative implication defined above, we add its converse to the knowledge base. 
For instance, the converse of the positive association in Eq. \eqref{eq:positive_implication} is defined as:
\begin{equation}
    y_k \implies \left( \bigwedge_{s \in \mathcal{A}_{k,l}} s \right).
    \label{eq:converse_implication}
\end{equation}
Converse rules for the negative associations are generated analogously.

\subsection{Phase 2: Concept-Coverage}
A stochastic sampling strategy may leave a subset of concepts in $S$ unselected, rendering them semantically ungrounded. 
To mitigate this, the second phase identifies all unused concepts and explicitly generates rules for them.
We repeat the logic of the Core Phase (generating positive, negative, and converse implications) for these specific concepts, ensuring that every $s \in S$ is grounded in at least one positive compositional rule within $\mathcal{R}$.

\section{Qualitative Results}
\label{sec:qualitatvie results}

\subsection{Positive Cases}
\label{ssec:positive case}
\begin{table*}[t]
    \centering
    
    \begin{tabular}{c c c}
        \includegraphics[width=0.3\textwidth]{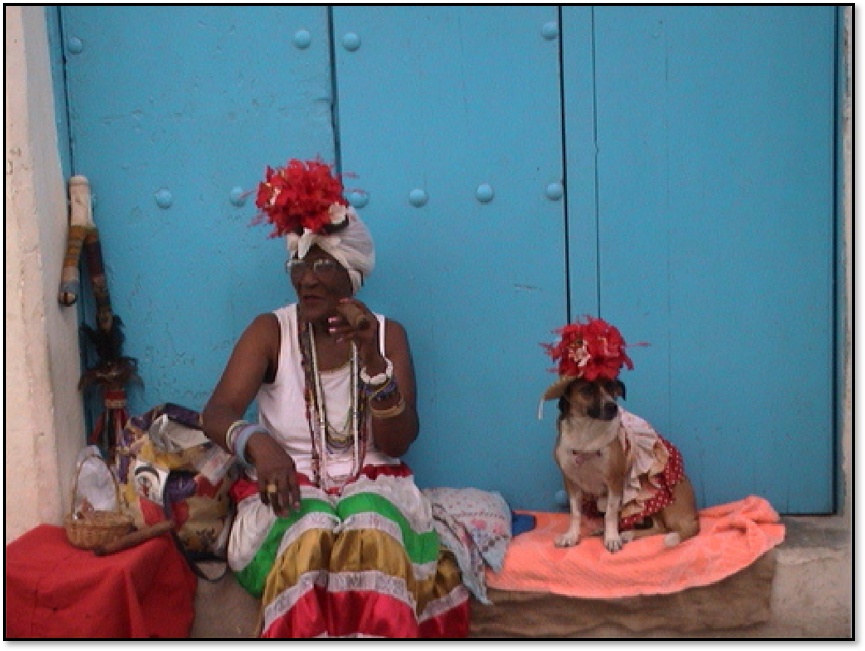} & 
        \includegraphics[width=0.3\textwidth]{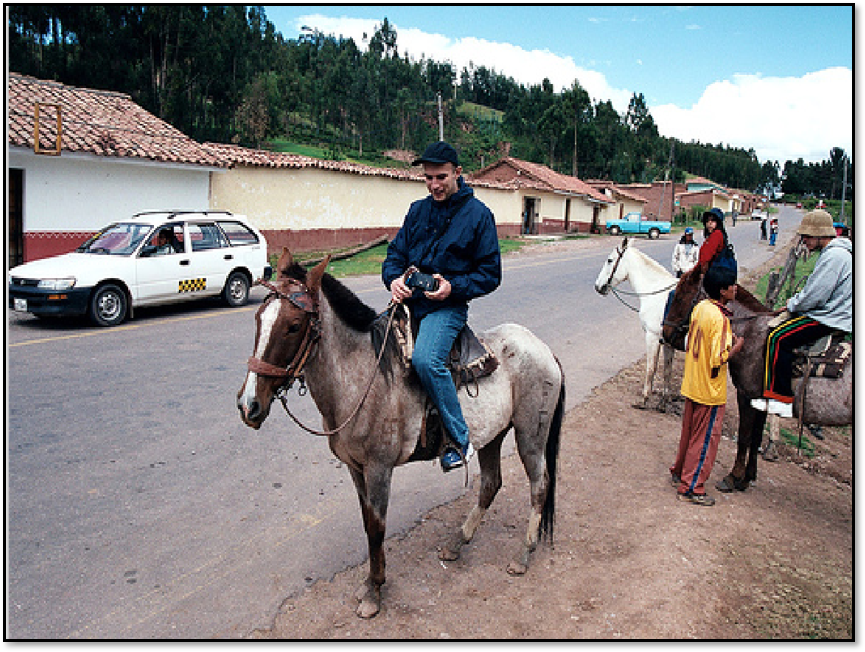} &
        \includegraphics[width=0.3\textwidth]{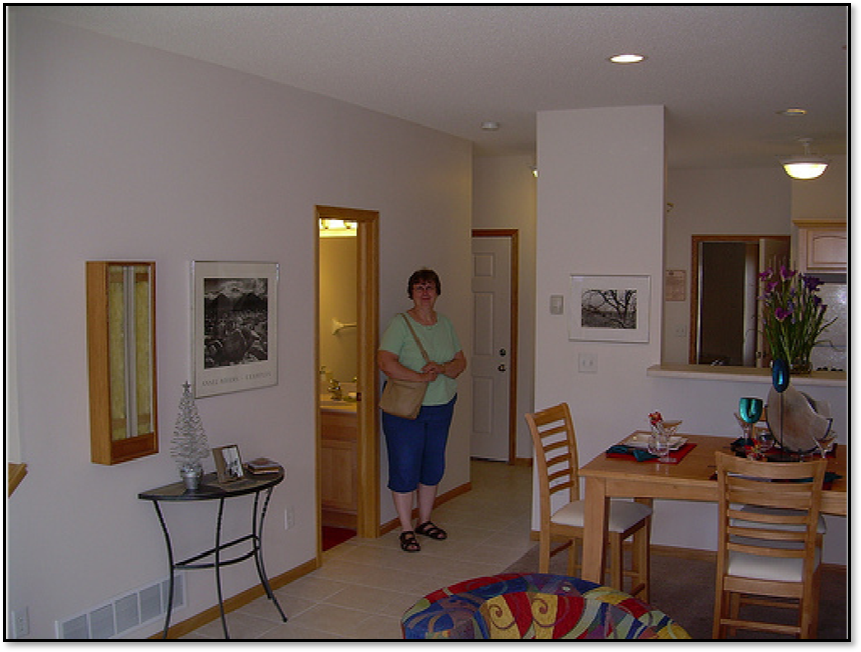} \\
        (a) & (b) & (c)
    \end{tabular}

    \small
    \setlength{\tabcolsep}{6pt}
    \renewcommand{\arraystretch}{1.4}
    \begin{adjustbox}{max width=\textwidth}
    \begin{tabular}{>{\centering\arraybackslash}c|c|c|c|ccc|ccc}
        \toprule
        \textbf{Image} &
        \textbf{Label Type} &
        \textbf{Label} &
        
        \multicolumn{1}{c|}{\textbf{WRN-101 (Baseline)}} &
        \multicolumn{3}{c|}{\textbf{KLUE}} &
        \multicolumn{3}{c}{\textbf{KLUE + $\mathcal{L}_{SAT}$}} \\
        
        &&& ${p_{cls}}^*$  & $p_{cls}$ & $p_\Delta$ & $p_\Delta -{p_{cls}}^*  $ & $p_{cls}$ & $p_\Delta$ & $p_\Delta -{p_{cls}}^*  $\\
        \midrule
        
        \multirow{3}{*}{(a)}
        & GT
        & person &
        0.81 &   0.82 & 0.99 \(\textcolor{green}{\uparrow +0.17}\) & \(\textcolor{green}{\uparrow +0.18}\) & 0.81 & 1.00 \(\textcolor{green}{\uparrow +0.19}\) &\(\textcolor{green}{\uparrow +0.19}\)\\
        
        & GT
        & dog &
        0.59 &  0.77 & 0.98 \(\textcolor{green}{\uparrow +0.21}\) & \(\textcolor{green}{\uparrow +0.39}\) & 0.59 & 0.91 \(\textcolor{green}{\uparrow +0.32}\) & \(\textcolor{green}{\uparrow +0.32}\)\\
        
        & \textcolor{gray}{Non-GT}
        & \textcolor{gray}{bottle} &
        \textcolor{gray}{0.37} & 
        \textcolor{gray}{0.44} & \textcolor{gray}{0.31} \(\textcolor{green}{\downarrow -0.13}\) & \(\textcolor{green}{\downarrow -0.06}\) &
        \textcolor{gray}{0.37} & \textcolor{gray}{0.00} \(\textcolor{green}{\downarrow -0.37}\) & \(\textcolor{green}{\downarrow -0.37}\)\\
        
        \midrule
        
        \multirow{4}{*}{(b)}
        & GT & person & 0.75 & 0.82 & 1.00 \(\textcolor{green}{\uparrow +0.18}\)& \(\textcolor{green}{\uparrow +0.25}\)  & 0.92& 1.00\(\textcolor{green}{\uparrow +0.08}\) & \(\textcolor{green}{\uparrow +0.25}\)\\
        & GT & car & 0.22 &   0.77 & 0.99\(\textcolor{green}{\uparrow +0.22}\) & \(\textcolor{green}{\uparrow +0.55}\) & 0.63 & 0.97\(\textcolor{green}{\uparrow +0.34}\) & \(\textcolor{green}{\uparrow +0.61}\) \\
        & GT & horse & 0.65 &  0.97 & 1.00 & \(\textcolor{green}{\uparrow +0.35}\)  &  0.85 & 1.00 \(\textcolor{green}{\uparrow +0.15}\) & \(\textcolor{green}{\uparrow +0.35}\)\\
        & \textcolor{gray}{Non-GT} & \textcolor{gray}{bottle} &
        \textcolor{gray}{0.04} & 
        \textcolor{gray}{0.34} & \textcolor{gray}{0.09}\(\textcolor{green}{\downarrow -0.25}\) &  \(\textcolor{red}{\uparrow -0.05}\) &
        \textcolor{gray}{0.28} & \textcolor{gray}{0.00} \(\textcolor{green}{\downarrow -0.28}\) & \(\textcolor{green}{\downarrow -0.04}\)\\
        
        \midrule
        
        \multirow{4}{*}{(c)}
        & GT & person & 0.36 &   0.99 & 1.00 & \(\textcolor{green}{\uparrow +0.64}\) & 0.92 & 1.00\(\textcolor{green}{\uparrow +0.08}\) & \(\textcolor{green}{\uparrow +0.64}\)\\
        & GT & chair & 0.59 &   0.99 & 1.00 &  \(\textcolor{green}{\uparrow +0.41}\)& 0.69 & 0.99\(\textcolor{green}{\uparrow +0.30}\) & \(\textcolor{green}{\uparrow +0.40}\)\\
        & GT & diningtable & 0.24 &  0.76 & 0.98\(\textcolor{green}{\uparrow +0.22}\) &  \(\textcolor{green}{\uparrow +0.74}\)& 0.61 & 0.95\(\textcolor{green}{\uparrow +0.34}\) & \(\textcolor{green}{\uparrow +0.71}\)\\
        & GT & pottedplant & 0.16 &   0.29 & 0.09\(\textcolor{red}{\downarrow -0.20}\) & \(\textcolor{red}{\downarrow -0.07}\) &0.55 & 0.79 \(\textcolor{green}{\uparrow +0.24}\) & \(\textcolor{green}{\uparrow +0.63}\)\\
        
        \bottomrule
    \end{tabular}
    \end{adjustbox}
    
    \caption{Comparison of Baseline WideResNet-101 (WRN-101), the WRN-101-KLUE variant, and WRN-101-KLUE+$\mathcal{L}_{SAT}$ trained with additional loss. All models are trained on PASCAL VOC train set. The top row displays input images (a–c) from PASCAL VOC Val set. We denote the baseline class probability as ${p_{cls}^*}$. For KLUE variants, we report the initial probability ($p_{cls}$) and the refined probability ($p_\Delta$). The values adjacent to arrows within the ($p_\Delta$) column represent the internal refinement magnitude ($p_\Delta - p_{cls}$). The separate column $p_\Delta -{p_{cls}}^*$ quantifies the absolute performance gain/loss of the final refined probability over the original baseline.  Rows denote Ground-Truth (GT) labels and Non-GT labels (shown in gray). Green and red text indicate positive and negative effects, respectively. For clarity, absolute difference values smaller than 0.05 (5\%) are omitted from the table.}
    \label{tab:pos_qualitative_results}
\end{table*}

Table~\ref{tab:pos_qualitative_results} highlights the qualitative benefits of integrating the KLUE bottleneck into the WideResNet-101 (WRN-101) architecture for multi-label classification. We compare the baseline WRN-101 against two KLUE variants: the standard KLUE and KLUE + $\mathcal{L}_{SAT}$. Both variants utilize WRN-101 as their backbone. The KLUE + $\mathcal{L}_{SAT}$ variant is trained with an additional loss term applied to a subset of rules $R_a$ (cf. Sec.~\ref{sec:Experiments} in the main paper). All models were hyperparameter-optimized using the Optuna framework on the PASCAL VOC Train set; general performance metrics are detailed in the main paper (cf. Table~\ref{tab:optuna_mAP_on_pascalvoc2012ChestMNIST}). For this qualitative evaluation, we selected the best-performing model from each category.
As shown in Table~\ref{tab:pos_qualitative_results}, KLUE variants demonstrate strong performance gains over the baseline. For example, in image (a), the best KLUE model improves the probability for the Ground-Truth (GT) class "dog" by approximately $32\%$, while simultaneously suppressing probabilities for the Non-GT "bottle" class. A similar trend is observed in images (b) and (c). Notably, in image (c), the KLUE + $\mathcal{L}_{SAT}$ variant yields a significant improvement for the "pottedplant" class. This suggests that the additional $\mathcal{L}_{SAT}$ term induces a bidirectional logical structure, providing better learning signals during optimization. This allows the model to learn meaningful implicit concepts, effectively treating the learned rules as targeted knowledge that enhances final classification performance.

\subsection{Failure Cases}
\label{ssec:failure case}
\begin{table*}[t]
    \centering
    
    \begin{tabular}{c c c}
        \includegraphics[width=0.3\textwidth]{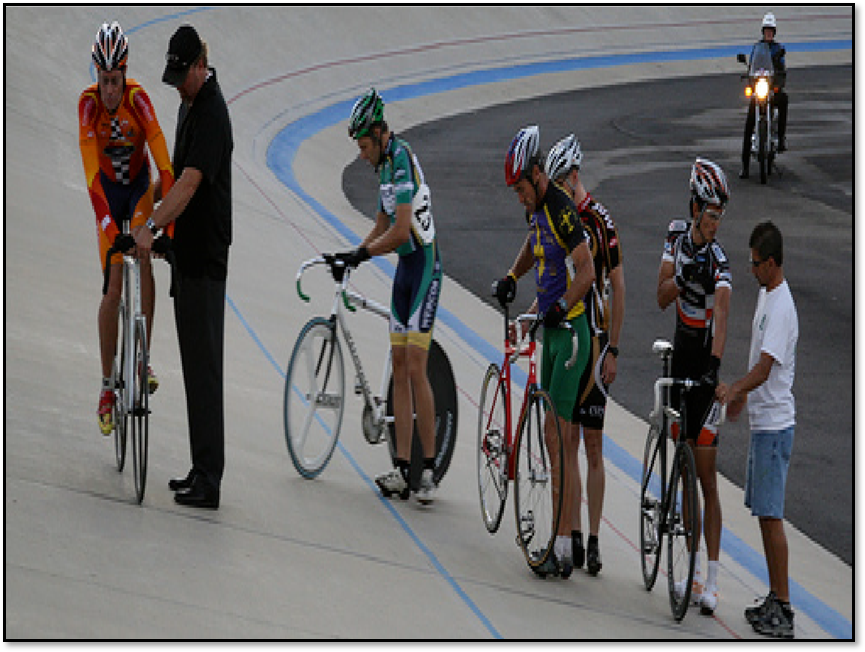} & 
        \includegraphics[width=0.3\textwidth]{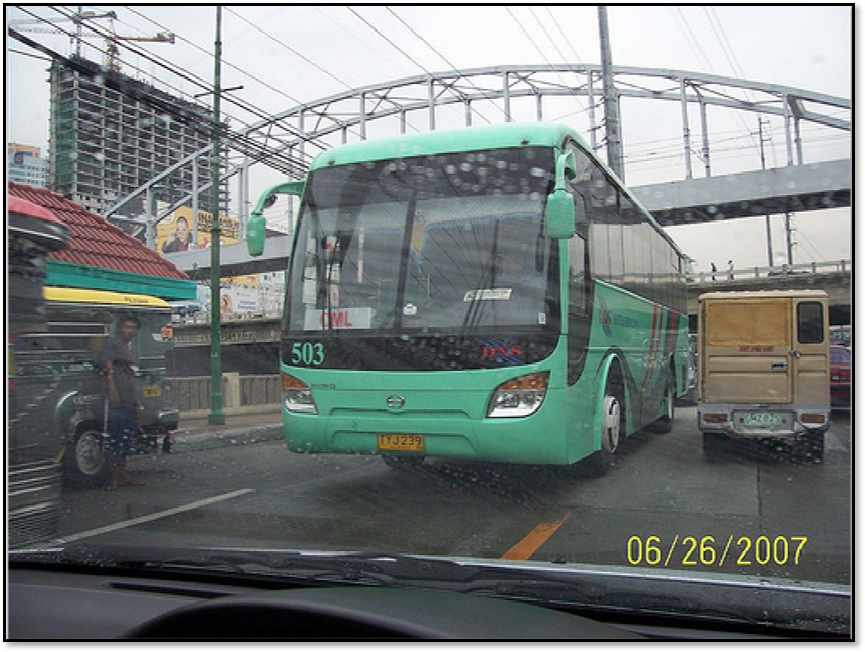} &
        \includegraphics[width=0.3\textwidth]{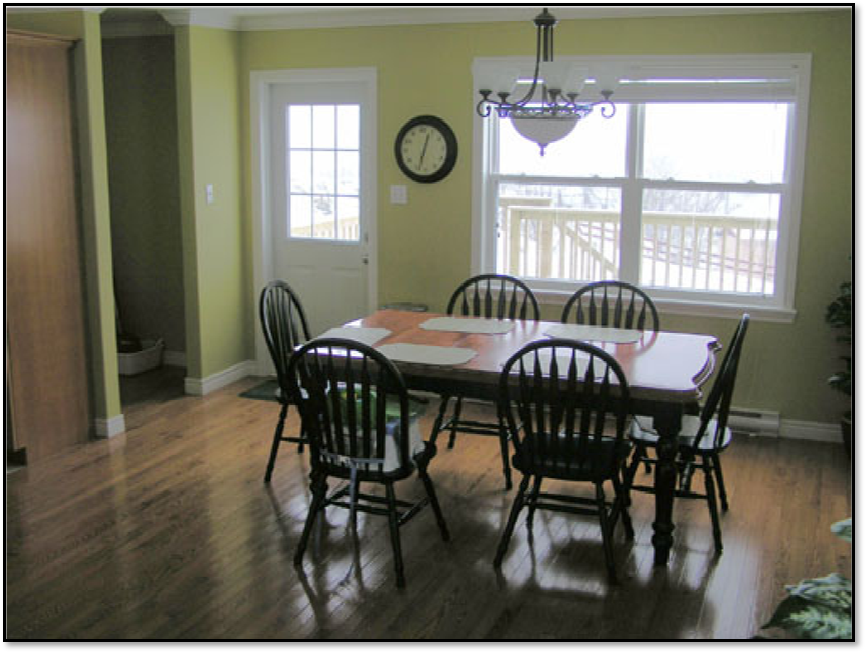} \\
        (d) & (e) & (f)
    \end{tabular}
    
    \vspace{1em} %
    
    \small
    \setlength{\tabcolsep}{6pt}
    \renewcommand{\arraystretch}{1.4}
    \begin{adjustbox}{max width=\textwidth}
    \begin{tabular}{>{\centering\arraybackslash}c|c|c|c|ccc|ccc}
        \toprule
        \textbf{Image} &
        \textbf{Label Type} &
        \textbf{Label} &
        
        \multicolumn{1}{c|}{\textbf{WRN-101 (Baseline)}} &
        \multicolumn{3}{c|}{\textbf{KLUE}} &
        \multicolumn{3}{c}{\textbf{KLUE + $\mathcal{L}_{SAT}$}} \\
        
        &&& ${p_{cls}}^*$  & $p_{cls}$ & $p_\Delta$ & $p_\Delta -{p_{cls}}^*  $ & $p_{cls}$ & $p_\Delta$ & $p_\Delta -{p_{cls}}^*  $\\
        \midrule
        
        \multirow{3}{*}{(d)}
        & GT
        & person &
        0.98  &  1.00 & 1.00  & \(\textcolor{green}{\uparrow +0.02}\) & 0.98 & 1.00 & \(\textcolor{green}{\uparrow +0.02}\) \\
        
        & GT
        & bicycle &
        0.89  & 1.00 & 1.00 & \(\textcolor{green}{\uparrow +0.11}\) &  0.99 & 1.00 & \(\textcolor{green}{\uparrow +0.11}\) \\
        
        & GT
        & \ul{motorbike} &
        0.18  & 0.05 & 0.00 \(\textcolor{red}{\downarrow -0.05}\)&  \(\textcolor{red}{\downarrow -0.18}\) &0.20 & 0.00\(\textcolor{red}{\downarrow -0.20}\) & \(\textcolor{red}{\downarrow -0.20}\)\\
        
        \midrule
        
        \multirow{3}{*}{(e)}
        & GT & \ul{person} & 0.30 &  0.32 & 0.06\(\textcolor{red}{\downarrow -0.26}\)& \(\textcolor{red}{\downarrow -0.30}\)&  0.45& 0.23\(\textcolor{red}{\downarrow -0.22}\)&\(\textcolor{red}{\downarrow -0.07}\) \\
        & GT & bus & 0.88  &  1.00 & 1.00 &\(\textcolor{green}{\uparrow +0.12}\) &  0.88 & 1.00\(\textcolor{green}{\uparrow +0.12}\) &\(\textcolor{green}{\uparrow +0.12}\) \\
        & GT & car & 0.62 &  0.97 & 1.00 & \(\textcolor{green}{\uparrow +0.38}\)& 0.85 & 1.00 \(\textcolor{green}{\uparrow +0.15}\) &\(\textcolor{green}{\uparrow +0.38}\)\\
        
        \midrule
        
        \multirow{3}{*}{(f)}
        & GT & chair & 0.82 &   1.00 & 1.00 & \(\textcolor{green}{\uparrow +0.18}\) &  0.96 & 1.00 & \(\textcolor{green}{\uparrow +0.18}\)\\
        & GT & diningtable & 0.32 &  0.85 & 1.00\(\textcolor{green}{\uparrow +0.15}\) & \(\textcolor{green}{\uparrow +0.68}\) & 0.82 & 0.99\(\textcolor{green}{\uparrow +0.17}\) &\(\textcolor{green}{\uparrow +0.67}\) \\
        & GT & \ul{pottedplant} & 0.18 &   0.19 & 0.00\(\textcolor{red}{\downarrow -0.19}\) & \(\textcolor{red}{\downarrow -0.18}\) & 0.35 & 0.02\(\textcolor{red}{\downarrow -0.33}\)&\(\textcolor{red}{\downarrow -0.16}\) \\
        
        \bottomrule

    \end{tabular}
    \end{adjustbox}
    
    \caption{Comparison of Baseline WideResNet-101 (WRN-101), the WRN-101-KLUE variant, and WRN-101-KLUE+$\mathcal{L}_{SAT}$ trained with additional loss. All models are trained on PASCAL VOC train set. The top row displays input images (d–f) from PASCAL VOC Val set. We denote the baseline class probability as ${p_{cls}^*}$. For KLUE variants, we report the initial probability ($p_{cls}$) and the refined probability ($p_\Delta$). The values adjacent to arrows within the ($p_\Delta$) column represent the internal refinement magnitude ($p_\Delta - p_{cls}$). The separate column $p_\Delta -{p_{cls}}^*$ quantifies the absolute performance gain/loss of the final refined probability over the original baseline.  Rows denote Ground-Truth (GT) labels, whereas green and red text indicate positive and negative effects, respectively. For clarity, absolute difference values smaller than 0.05 (5\%) are omitted from the table.}
    \label{tab:failure_cases}

\end{table*}

Table~\ref{tab:failure_cases} presents an evaluation of the models on "hard" examples where all the models struggles. We observe that in certain scenarios, the KLUE variants are unable to correct the predictions.
For instance, in image (d), the KLUE models fail to improve the probability for the "motorbike" class. 
Similarly, in image (e), the probability for the "person" class remains low, likely due to the visual ambiguity of the subject in the scene; however, it is worth noting that the model significantly improves confidence for the "car" class in the same image.
A similar limitation is seen in image (f) for the "pottedplant" class. 
We attribute this to partial visibility, where the learned concepts fail to sufficiently validate the class presence to refine the probability.

\textbf{Conclusion}. In general, KLUE shows limitations when dealing with heavy occlusion or partial visibility in a multi-label classification setting.
However, we hypothesize that this could be overcome in an object detection framework. In detection tasks, the model receives additional learning signals from the object's spatial location, which KLUE could leverage to direct its attention more effectively. 
Validating this hypothesis remains a promising direction for future work.

The corresponding tables for these assessments are presented on the subsequent pages.

\subsection{Activation Comparison}

\begin{figure}[h]
  \centering
  \includegraphics[width=\linewidth]{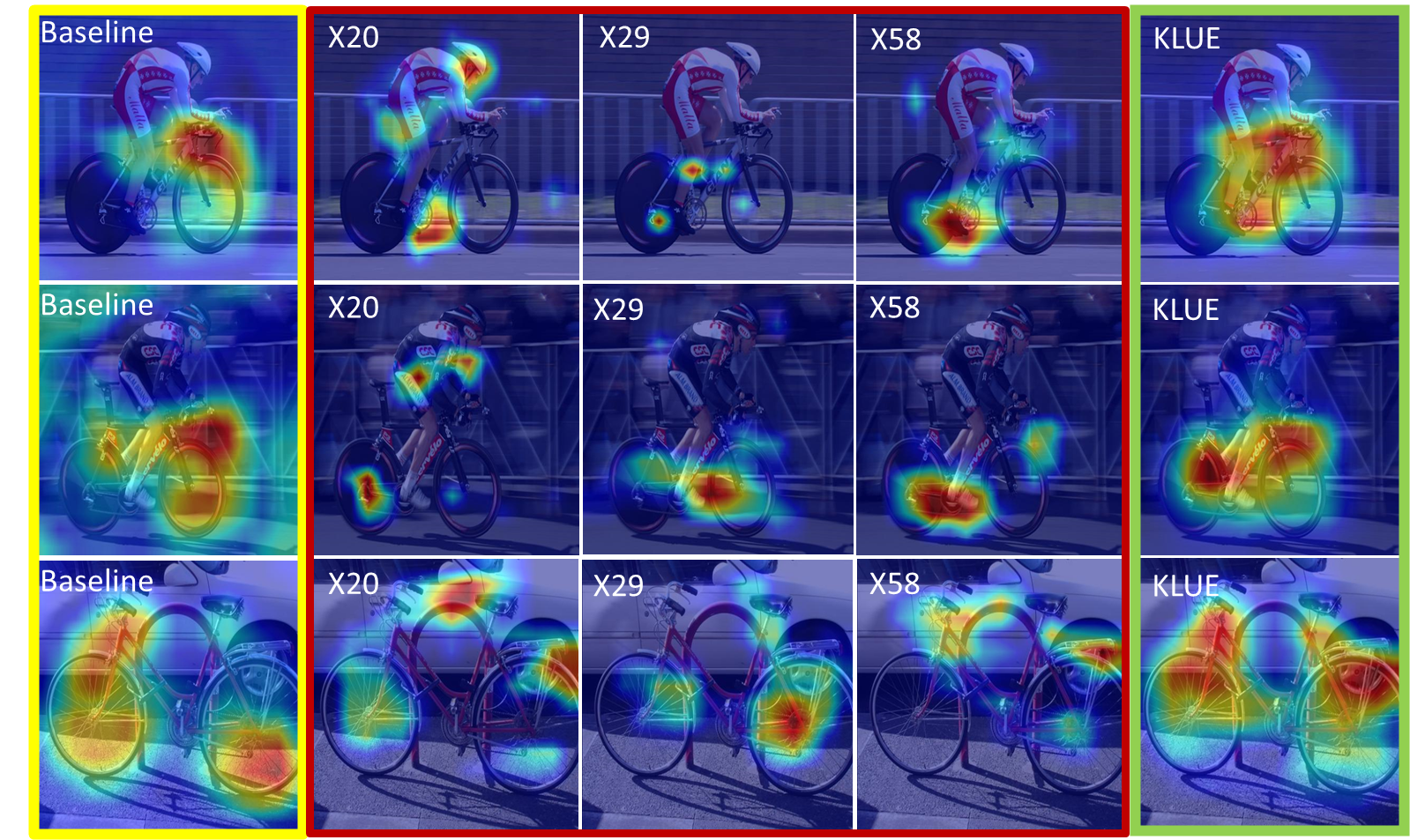}
  \vspace{-0.0mm}
  \caption{\small Activation comparison: Baseline WRN-101 (yellow), KLUE concepts (red), KLUE final (green).}
  \vspace{-0.0mm}
  \label{fig:rebuttal_fig_activations}
\end{figure}
Fig.~\ref{fig:rebuttal_fig_activations}, recurring concepts (e.g., x20/x29/x58) activate consistently for semantically similar images and differ for a distinct one; compared to the baseline, KLUE yields more localized, refined activations.

\subsection{Scalability / Overhead} 
\begin{figure}[h]
  \centering
  \includegraphics[width=0.485\columnwidth]{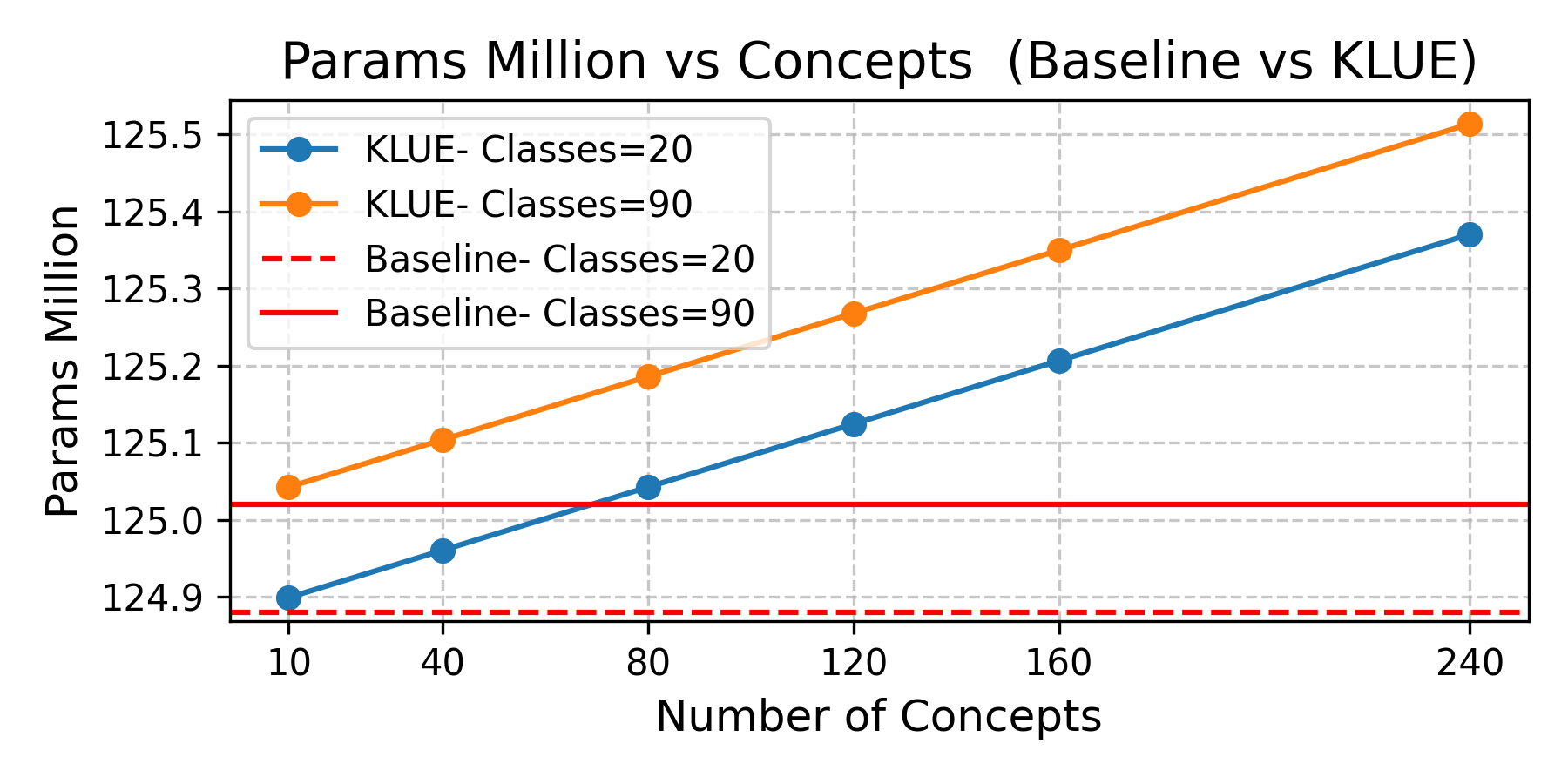}
  \hfill
  \includegraphics[width=0.485\columnwidth]{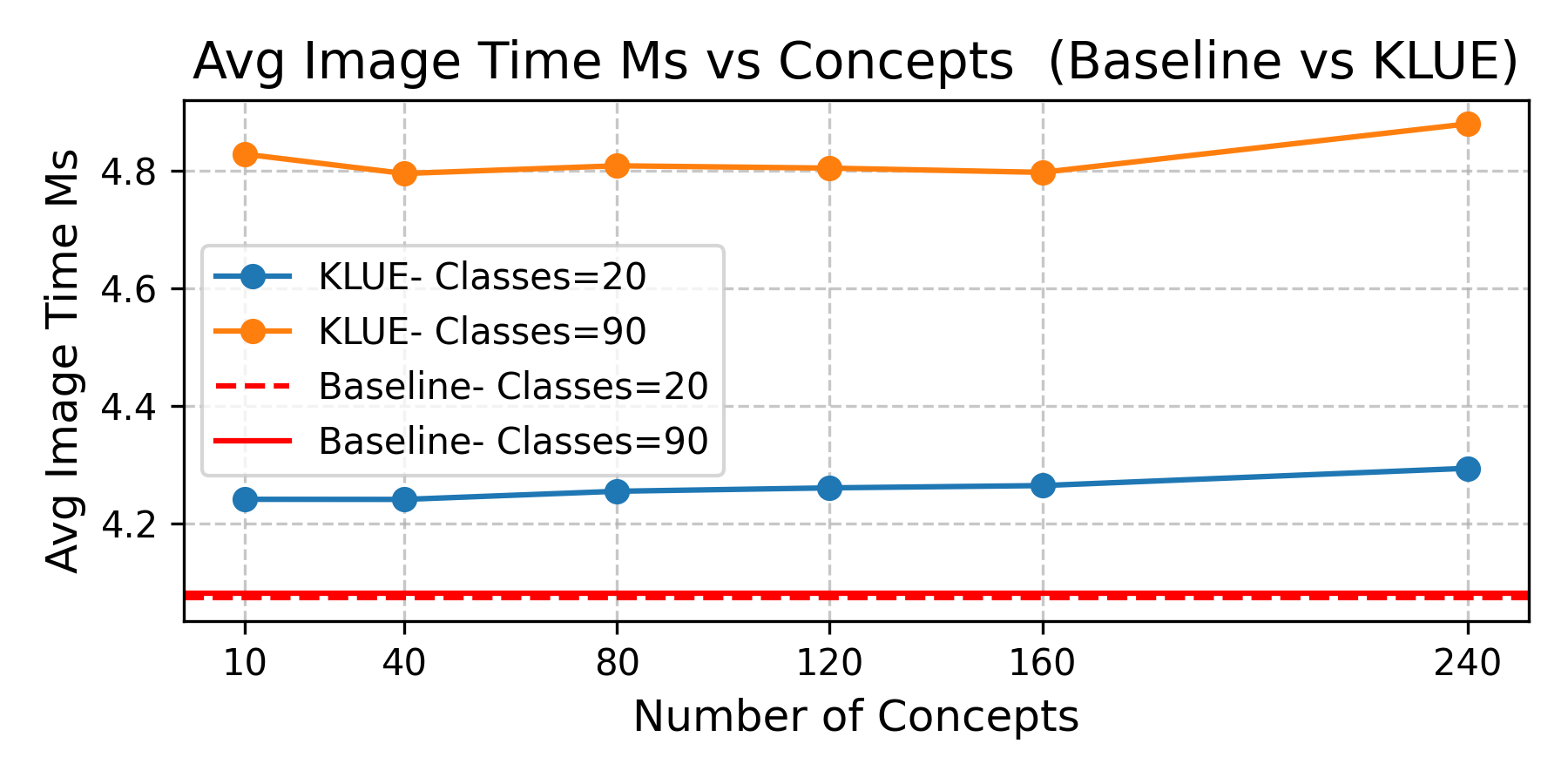}
  \vspace{-2.0mm}
  \caption{\small Minimal parameter overhead (left) and constant latency (right) across concept scales. Backbone: WRN-101}
  \vspace{-3.5mm}
  \label{fig:scalability}
\end{figure}
Fig.~\ref{fig:scalability} shows KLUE is lightweight: $<0.5\%$ parameter overhead as classes/concepts grow and \textbf{nearly constant inference latency} from 10--240 concepts.

\end{document}